# Detection of Non-Stationary Photometric Perturbations on Projection Screens

Miguel Castañeda-Garay[1], Óscar Belmonte-Fernández[2],
Hebert Pérez-Rosés[3], Antonio Díaz Tula[1]

[1] Department of Computer Science,
University of Oriente, Santiago de Cuba, Cuba
[2] Department of Languages and Information Systems,
University Jaume I, Castellón, Spain
[3] School of Electrical Engineering and Computer Science
The University of Newcastle, Australia

**Abstract.** Interfaces based on projection screens have become increasingly more popular in recent years, mainly due to the large screen size and resolution that they provide, as well as their stereo-vision capabilities. This work shows a local method for real-time detection of non-stationary photometric perturbations in projected images by means of computer vision techniques. The method is based on the computation of differences between the images in the projector's frame buffer and the corresponding images on the projection screen observed by the camera. It is robust under spatial variations in the intensity of light emitted by the projector on the projection surface and also robust under stationary photometric perturbations caused by external factors. Moreover, we describe the experiments carried out to show the reliability of the method.

## 1 Introduction

Large projection screens have become very popular nowadays in interactive systems. LCD monitor panels and projector arrays are being used for that purpose, resulting in unified high-resolution screens, in some cases with stereo visualization. Those screens enable the presence of multiple viewers, detailed rendering of models, sensation of immersion, and a natural collaborative environment among several users.

Screens made up of projector arrays show some geometric and photometric anomalies between adjacent tiles, as well as inside each tile, causing notable irregularities at their unions. The irregularities comprise mismatching between adjacent projected images, shape and size distortions, and photometric differences at the tile borders due to the effects of light superposition and dispersion. Those anomalies have given rise to a new challenge: the creation of seamless tiled display. There are several works devoted to that goal, which essentially carry out a dual calibration process: geometric and photometric calibration. Some details about the construction of such screen can be found in Bresnahan et al. (2003), Kresse, Reiners and Knöpfle (2003), and Brown, Majumder and Yang (2005).

Moreover, in the projection-based systems images are formed on the projection screen from the light emitted by projectors onto that surface. Therefore, the surrounding light, obstacles in the way of light beams, and irregularities on the projection surface, among other factors, affect image visualization. In that sense, several applications based on computer vision techniques have been developed to detect and correct some photometric and geometric perturbations on the projection surface. These applications include, for instance, those for detecting and suppressing the shadow created by occluding users in front projection systems (Rehg, Flagg, Chan, Sukthankar and Sukthankar, 2002, and Jaynes, Webb and Steele, 2004).

The work presented here addresses problems of the same nature. It starts with a modification of the global method developed by Jaynes et al. (2004) to detect shadows in projected images, and ends with a local method to detect photometric perturbations in projected images.

The method presented in Jaynes et al. (2004) is quite effective for the detection of shadows, and can be used in general to detect photometric perturbations in projected images in real time. It employs several cameras and projectors located in front of the screen, and relies on the comparison between the frames stored in the frame buffer of the projectors and the corresponding frames captured by the cameras. In order to perform that comparison, a preliminary photometric calibration process among projectors, cameras, and the screen is needed. The calibration method gives rise to transfer functions between colour spaces, which enable to estimate the images to be captured by the cameras, from the frame buffers images of projectors. Then, a comparison is made between the images observed by the cameras, and the corresponding estimated images. In that work, a global photometric calibration is performed, so that all the regions of the screen are processed uniformly. Therefore, the method does not consider the existence of locally stationary photometric perturbations, that can be due to intrinsic or extrinsic factors, which makes the system to lose some accuracy during detection.

Among the intrinsic factors we could cite variations in the spatial intensity of the projected beams on the projection surface, caused by the location of the projectors and the specific properties of the screen's material. The extrinsic factors include the incidence or lack of external light on some projection areas, and irregularities or stains on the screen.

As for other shortcomings of the method presented in Jaynes et al. (2004), we could cite in general the lack of synchronization between projectors and cameras, the high latency, and the global character of photometric calibration.

In this work we give a local method for real-time detection of local photometric perturbations in projected images, which is an improvement over the method presented in Jaynes et al. (2004). Our goal is to improve the accuracy and reliability of photometric perturbation detection in the presence of spatial variations in light intensity on the projection surface, and the influence of locally stationary external factors. Our field of action lies in the calibration process among the cameras, the projection screen, and the projectors. For the experiments, we have developed a prototype that uses only one camera/projector pair,[5] as shown in Figure 1.

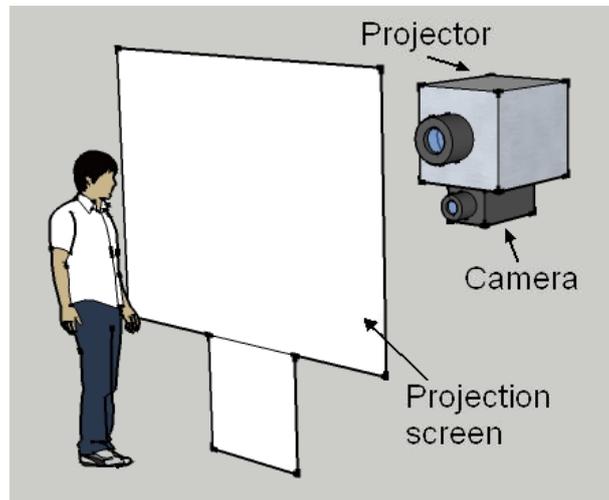

Figure 1. Setting of projector, camera, and projection screen.

Moreover, in the work we provide all implementation details of the proposed method, and a set of experiments is described, that show the superiority of our method over that of Jaynes et al. (2004) in terms of accuracy and reliability.

These results can be applied to the construction of an interactive interface, based on a projection screen. A potential scenario is a lecture room, where the camera-projector pair is located in the ceiling or the back of the room, and the projection screen is on the stage. The lecturer stands behind the screen (as in Figure 1), and interacts with the system by touching the back of the screen. Thus, the lecturer does not interfere with the students' field of vision, and can push different buttons, turn pages, etc., more or less in the same manner as with a touch screen.

The paper is structured as follows. In Section 2 we give a synthesis of the previous work related with the subject. In Sections 3 and 4 we provide our respective models for geometric and photometric calibration of camera, projector and screen. In Section 5 we describe the processes of image segmentation and comparison, for detecting the regions that have suffered some photometric perturbation. In Sections 6 and 7 we give the details for an efficient implementation of the proposed method. In particular, Section 6 is devoted to general implementation issues, while Section 7 describes a parallel GPU-based implementation. In Section 8 we describe the prototype of the system that has been developed for experimentation purposes. In Section 9 we describe the experiments carried out, and the results obtained, that confirm the greater accuracy and reliability of our method in comparison with the one presented by

---

[5] Having multiple cameras does not imply any change in the system's architecture, and it might help to increase the robustness of the system, but that needs to be confirmed in practice.

Jaynes et al. (2004). Finally, in Section 10 we present our conclusions, and guidelines for future work.

## 2   Previous work

A user's interface for interacting with a projection screen with the aid of a laser pointer is shown in Kirstein and Muller (1998). To detect the laser spot on the screen, a reference image is systematically computed from the last frames captured. The difference between the current frame and the reference image is employed to determine the regions that have suffered photometric variations.

In Sukthankar, Stockton and Mullin (2000), a user's interface to interact with a projection screen is described. In that interface, the user can control the presentation in various ways, which include moving a laser pointer, or finger interaction. The difference between the last image captured and the previous one is used to detect the regions that have undergone some photometric variation, among other methods.

In Sukthankar, Stockton and Mullin (2001), a method for geometric calibration between camera and projector is described. That method is based on the computation of a planar homography between the projected image and the corresponding image captured by a camera. The homography establishes a correspondence between the frame buffer coordinates of the projector and the camera, respectively.

Some works have employed threshold segmentation to detect laser spots on projected images, taking into account that the laser spot constitutes a region of high brightness (Eckert and Moore, 2000, Olsen and Nielsen, 2001, Lapointe and Godin, 2005, Ahlborn, Thompson, and Krelylos, 2005). The work presented in Ahlborn et al. (2005) shows a local method to improve detection of the laser spot in the presence of spatial and temporal light intensity anomalies on the screen.

In Jaynes et al. (2004), a method is shown to detect shadows on frontal projection systems with several projectors, also based on image comparison. In this method, a photometric calibration model is used to estimate the images to be captured by the cameras from the images contained in the frame buffers of the projectors. Based on that, and on the comparison between real and estimated images, photometric alterations undergone by the projected images can be determined.

In this work we give a local method for real-time detection of non-stationary photometric perturbations in projected images, based on computer vision techniques. We use the method described in Sukthankar et al. (2001) to establish the geometric calibration between camera, screen, and projector. Then, the global method for detecting photometric perturbations employed in Jaynes et al. (2004) is modified, thus obtaining a local method with improved reliability for detecting non-stationary photometric perturbations on the screen. Part of this work was presented at the Spanish Congress of Graphics 2009 (Castañeda-Garay, Belmonte-Fernández, Gil-Altaba, Pérez-Rosés and Coma, 2009).

## 3 Geometric calibration of camera and projector

In order to establish the coordinate correspondence between the camera images, the screen images, and the projector's frame buffer images, a geometric calibration process is necessary. The result of that process is a set of transfer functions between the respective coordinate systems.

For that purpose, the camera should be oriented in such a way that its field of vision totally contains the projected images. We define the *region of interest* of the camera as the region that contains exclusively the valid images on the screen. The extra area seen by the camera will be called *periphery*. The points belonging to the periphery are invalid points of the camera frames, in the sense that they do not correspond to any point in the projector's frame buffer.

Geometric calibration establishes a correspondence between the coordinate systems of the estimated images and the images taken by the camera. The purpose of geometric calibration is twofold. First, it is necessary for photometric calibration of camera, projector, and screen. Second, it helps to reduce the computational cost of detecting the areas subject to photometric perturbations, since detection can focus on the region of interest only.

In order to establish this correspondence we have employed the method of planar homography described in Sukthankar et al. (2001). The correspondence of a point ($x$, $y$) in the projector's frame buffer with any point on the projection screen, and consequently with a point ($X$, $Y$) in the camera's frame buffer, can be described by a projective transformation of the form:

$$(x, y) = \left( \frac{p_1 X + p_2 Y + p_3}{p_7 X + p_8 Y + p_9}, \frac{p_4 X + p_5 Y + p_6}{p_7 X + p_8 Y + p_9} \right) . \tag{1}$$

Four pairs of corresponding points, such that no three of them are collinear, are enough to estimate the parameters $p_1, \ldots, p_9$. For more details see Sukthankar et al. (2001). For that purpose we use the four corners of the region of interest.

In order to detect the aforementioned corners, a blank image is first projected and captured by the camera. Then, the four border lines of the blank image are calculated with the aid of a global-automatic-threshold segmentation technique, boundary-extraction morphological operations, and the Hough transform (Gonzalez and Woods, 2002). Finally, the four corner points are obtained as the intersections of the border lines.

The domain of the transfer functions is defined in the coordinate system of the estimated image buffer, and the co-domain in the coordinate system of the camera image buffer. The following equations describe these transfer functions:

$$H^x(X,Y) = \frac{p_1 X + p_2 Y + p_3}{p_7 X + p_8 Y + p_9} . \tag{2}$$

$$H^y(X,Y) = \frac{p_4 X + p_5 Y + p_6}{p_7 X + p_8 Y + p_9} \quad . \tag{3}$$

where $X$ and $Y$ represent, respectively, the abscissa and the ordinate of a point in the coordinate system of the estimated image buffer, $H^x(X,Y)$ and $H^y(X,Y)$ represent the coordinates of the corresponding point in the camera's image buffer, and the parameters $p_1, \ldots, p_9$, represent the homography coefficients, calculated from the correspondence between the four corner points of the region of interest and the four corner points of the buffered estimated images.

## 4   Photometric calibration of camera and projector

The photometric perturbations on a projection screen can be determined with the aid of comparisons between the projector's frame buffer images, and the corresponding images taken by a camera. However, this process should not be carried out directly, since the projector's frame buffer images and the images taken by the camera have different photometric spectra, in general. That difference is due to several internal and external factors, such as the difference in colour spectra and brightness levels of camera and projector, the location of the projector with respect to the screen, which causes variations in brightness, the photometric properties of the projection screen's material (e.g. its reflectance and dispersion properties), the influence of ambient light on the screen, as well as the particular features of the camera, the adjustment of the intrinsic and extrinsic parameters, and the presence of irregularities or spots on the projection surface, among other factors. Hence, in order to perform the comparison effectively, it is necessary to previously correct the stationary photometric alterations undergone by images after being projected and captured by the camera.

In Jaynes et al. (2004) a work was presented to model that situation, where a colour model was used for each camera/projector pair used. Several experiments made with different camera/projector pairs and different projection surfaces have suggested Verhulst functions as candidate transfer functions to estimate the images to be captured by the camera from the images in the projector's frame buffer. For each camera/projector pair, and for each colour component (red, green, blue), a Verhulst function is estimated, with the form:

$$F_c(z) = \frac{a}{1 + e^{-a(z-b)}} + k \quad . \tag{4}$$

For each camera/projector pair, for each colour component *c*, and for each intensity value *z* of colour *c* in the projector's frame buffer, the function $F_c(z)$ yields an estimation of the intensity value to be read by the camera.

The parameters *a*, *k*, α, and *b*, of Equation 4, are estimated separately for each colour component, thus giving rise to three transfer functions. In order to estimate the parameters are projected four sample images with uniform intensity at an early configuration stage. These images are observed by the camera, and the average intensity is calculated. Then, these four pairs of corresponding intensities are used to compute the parameters. For more information see Jaynes et al. (2004).

Nevertheless, in spite of the adequate results obtained with this model, it still has the disadvantage of being global. During the estimation of the parameters of Equation 4, the intensities observed by the camera are averaged, which results in the loss of local information. Hence, the photometric calibration process is applied uniformly to all areas of the projector's frame buffer images, and consequently, to all areas of the projected and captured images. Therefore, this method should be sensitive to spatial variations of beam intensity on the projection surface, and to the influence of external locally stationary factors on the screen as well.

Therefore, we propose a local modelling method. To this end, the projector's frame buffer is partitioned into regions, and the transfer functions are specific to those regions.

To model this situation, the projector's frame buffer is conceived as a rectangular grid or matrix, made up of disjoint cells, whose union is equal to the whole buffer. The number of rows and columns is chosen so as to match the camera resolution, which have the same resolution that the estimated image. For the sake of reducing the latency time, we take the camera resolution to be smaller than the resolution of the projector's frame buffer. Therefore, each region of the projector's frame buffer corresponds to a point in the estimated image buffer. Formally, we define the region $R_{xy}$ of the projector frame buffer as:

$$R_{xy} = \left\{ (x', y') \in B \Big/ \left[ \frac{x' \cdot |E|^x}{|B|^x} \right] = x, \left[ \frac{y' \cdot |E|^y}{|B|^y} \right] = y \right\} . \tag{5}$$

where *B* represents the projector's frame buffer, $|B|^x$ and $|B|^y$ represent the number of columns and rows of *B*, respectively, and $|E|^x$ and $|E|^y$ represent the number of columns and rows of the estimated image buffer.

With this partition, for each colour component and each region, a different transfer function is fitted. In other words, for each colour component and each buffer region, we estimate different sets of parameters *a*, *k*, α, and *b*, one set for each colour transfer function. This results in a local method that depends not only on the colour components and the pixel intensities in the projector's frame buffer, but also on the particular region in the projector's buffer, and on its corresponding region in the camera's buffer as well. We obtain for each colour component *c* a family of colour transfer functions of the form:

$$F_{xy}^c(z) = \frac{a_{xy}^c}{1+e^{-a_{xy}^c(z-b_{xy}^c)}} + k_{xy}^c \quad (6)$$

For colour component *c*, this equation represents the colour transfer function of the region $R_{xy}$ of the projector's frame buffer, and $a_{xy}^c$, $\alpha_{xy}^c$, $b_{xy}^c$ and $k_{xy}^c$ are its respective parameters. With this approach we can improve the effectiveness of detection of photometric perturbations on the projected images, since each transfer function is fitted for a specific region.

The estimated images have the same resolution that the camera images taking into account that we will have to compare them. The prototype that has been implemented operates with a maxima resolution of the camera of 640 x 480 pixels. The projector's frame buffer resolution is 1024 x 768 pixels.

In Figure 2 we show the correspondence that is established between a projector's frame buffer region, a pixel in the estimated image buffer, and a pixel in the buffer of the camera. To each region of the projector's frame buffer corresponds exactly one proper point in the estimated image buffer. In turn, to each proper point in the estimated image corresponds exactly one proper point in the image buffer of the camera. However, to a proper point in the camera's buffer can correspond 0, 1, or more proper points of the estimated image buffer, since the region of interest in the camera's image buffer is smaller than the buffer itself.

In this sense, besides fitting the aforementioned transfer functions, we have to model the effect undergone by the camera when it captures an image of smaller resolution than the one that is projected. Here we have used the average intensity of the pixels belonging to the region $R_{xy}$, in the projector's frame buffer, as the intensity value for pixel (*x*, *y*) of the estimated image.

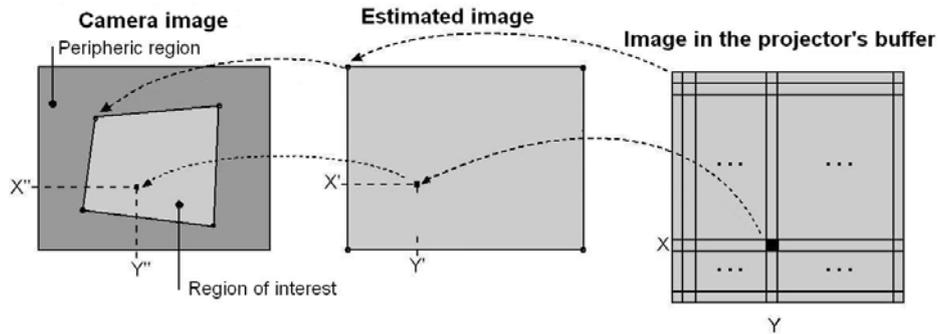

Figure 2. Correspondence between a region of the projector's frame buffer, a point in the estimated image buffer, and a point in the camera's image buffer.

In order to get a camera model that is as realistic as possible, before computing the intensity of pixel (*x*, *y*) in the estimated image for some colour component, we first apply the colour transfer function corresponding to the region $R_{xy}$ of the projector's frame buffer (according to Equation 6) to each pixel within that region, and the resulting values are then averaged. This emulates the process where the information contained in all the pixels of region $R_{xy}$ is captured by the camera, and then processed to obtain a single pixel.

That way, for colour component *c*, the intensity of point (*x*, *y*) of the estimated image *E*, can be expressed as a function of the intensities of the points of region $R_{xy}$ of the projector's frame buffer *B*, as follows:

$$E_{xy}^c = \frac{\sum_{p \in R_{xy}} F_{xy}^c(B_p^c)}{|R_{xy}|} \quad . \tag{7}$$

where $B_p^c$ is the intensity of point *p* in the projector's frame buffer *B*, in colour component *c*, and $|R_{xy}|$ is the number of points contained in $R_{xy}$.

After this photometric calibration process, local photometric perturbations can be determined more effectively by comparing the images observed by the camera with the corresponding estimated images. In the next section we explain the comparison methodology.

## 5   Image segmentation and comparison

Since our region of interest in the camera's image buffer is smaller than the buffer itself, a point (*x*, *y*) of the buffer does not necessarily correspond to the same point (*x*, *y*) in the estimated image buffer. Therefore, in order to compare the observed image with the estimated image, we must first find the correspondence between the pixels of each image.

In order to find that correspondence, we have used here the method of planar homography described in Section 3. The comparison between the intensity levels of two pixels reduces to the difference of their absolute values. During the segmentation process, that difference is then compared in turn with a global threshold.

More formally, the difference in absolute value for every pixel $E_{xy}$ of the estimated image and its corresponding pixel $C_{x'y'}$ of the observed image is compared with a threshold *U* for every colour component, and this comparison yields the value 1 if for at least one colour component the intensity difference is greater than *U*, and 0 otherwise. For a specific colour component, this calculation is modelled by the following equation:

$$O_{xy} = \begin{cases} 1, if \left| E_{xy} - C_{x'y'} \right| > U \\ 0, otherwise \end{cases} \qquad (8)$$

where $x' = H^x(x,y)$, $y' = H^y(x,y)$, and $H^x$ and $H^y$ are defined according to Equations 2 and 3, respectively.

This yields a binary image *O*, with value 1 in all the regions affected by photometric perturbations, and value 0 in the remaining areas. The coordinates of the resulting points refer to the coordinate system of the estimated images, and the periphery of the camera images is discarded. Thus, a simple detection algorithm can be used to determine the coordinates of the areas affected by photometric perturbations, without worrying about the periphery. The algorithm works by detecting clusters of points with value 1.

## 6  Implementation of the method

The current version of the implementation uses a non self-calibrated video camera that captures the frames in RGB format, with a maximal resolution of 640 × 480 pixels, and a projector with a resolution of 1024 × 768 pixels. The implementation was carried out in Java, using the Java Media Framework for capturing and processing the frames.

The image was divided into 320 × 240 = 76800 regions for the computation of the colour transfer function. The size of the regions is chosen so as to accommodate the expected perturbations (in our case, the shadow of a finger touching the back side of the screen).

Real-time processing of the frames is a challenging task, especially in what concerns the evaluation of the transfer functions for each pixel and each colour component in the projector's frame buffer, and the evaluation of Equations 2 and 3 on each point of the estimated images. The best solution we have found for that purpose is to use pre-calculated tables that are created in an initial stage of the system's configuration.

The evaluation of Equations 2 and 3 for each buffer point of the estimated images, in order to obtain the corresponding point in the camera buffer, is only performed once during the system's configuration stage, constructing a table *H* of correspondences between the buffer points, of the form:

$$H_{xy} = (H^x(x,y), H^y(x,y)) = (x', y') \; . \qquad (9)$$

Then, during the detection phase of photometric perturbations, in order to establish a comparison between the estimated image and the camera image, for every pixel (*x*,

*y*) of the estimated image we can obtain the corresponding point (*x′*, *y′*) in the camera image without evaluating Equations 2 and 3, since $H_{xy} = (x', y')$.

The other costly process is the evaluation of colour transfer functions described by Equation 6 on the pixels of every colour component in the projector's frame buffer. This has to be done for every frame captured by the camera. In order to reduce the computational cost, we build three 3D tables during the system's configuration phase, one for each colour component, and they are pre-evaluated on each region $R_{xy}$ of the projector's frame buffer, for intensity values that are multiples of 4. The tables are constructed as follows:

$$E^c_{xyz} = F^c_{xy}(4 * z) . \qquad (10)$$

where $F^c_{xy}$ is defined according to Equation 6, and *z* takes values in the range 0..63. This leads to a four-time reduction in memory use, at the expense of a three-level increase in the estimation error.

Then during photometric perturbation detection, the evaluation of Equation 6 for an intensity value *z*, at a point of region $R_{xy}$, and for colour component *c*, can be obtained according to $E^c_{xyw}$, where $w = \left\lfloor \dfrac{z}{4} \right\rfloor$.

## 7 Parallelization of the estimation and comparison algorithms with CUDA.

The process of computing the estimated image and comparing this with the camera image can be parallelized using the Single Instruction Multiple Data (SIMD) model (Flynn, 1972).

Data parallelism can be implemented with the aid of the Compute Unified Device Architecture (CUDA) parallel computing technology. This cutting-edge technology demonstrates the enormous parallel computing capabilities of the NVIDIA's GPUs in a relatively easy-to-use environment for those applications that can be modelled by an SIMD machine.

It is important to note that the projector's frame buffer is located in Video RAM, as well as the CUDA's memory space, which implies that, for the estimation algorithm, the projector's frame buffer doesn't need to be copied into conventional RAM, thus avoiding the latency time, and traffic over the PCI-Express bus.

To make the projector's frame buffer available to the CUDA's threads for the estimation, an OpenGL's Pixel Buffer Object (PBO) is used. This PBO receives a copy of the projector's frame buffer by OpenGL's API functions. Then it is mapped using CUDA's API and passed to the threads as a pointer.

We used the output data decomposition technique for both the estimation and comparison algorithms, where each output element can be independently computed as a function of the input. The value of each pixel in the estimated image (output) depends only on the corresponding frame buffer's region and its pre-calculated colour tables (input). The value of each pixel in the binary image $O$ (output, see Equation 8) depends only on the corresponding pixels in both the estimated and camera images (input).

For the estimation algorithm, a task is defined as computing a pixel of the estimated image; for the comparison algorithm, a task is defined as computing a pixel of the resulting binary image $O$. Those tasks are agglomerated in such a way that a CUDA thread will carry out a certain number of tasks, resulting in a balance between the number of thread blocks and the multiprocessor count of the GPU.

The thread blocks will have $16 \times 16 = 256$ threads, and the size of the block grid depends on the camera's image dimensions.

Since each thread will access a different pre-calculated table entry during the estimation, and each table entry has 64 bytes, the memory access pattern will not be coalescent, thus degrading performance by introducing a significant latency time.

To counteract this fact, a shared memory matrix is used as a cache for the pre-calculated tables. By using a 32-bit word per thread for each colour component, we obtain a bank-conflict free shared memory access, which allows estimating up to 16 different colour intensities for each colour component without accessing the global memory.

This implementation was integrated with Java through the Java Native Interface. In the experimentation, using a camera resolution of $320 \times 240$ pixels, a projector's frame buffer resolution of $1024 \times 768$, a Core 2 Duo processor at 2.66 GHz, and a NVIDIA GeForce 8400 GS having two multiprocessors, the average latency time was 28 milliseconds. This allows to process up to 35 frames per second.

## 8 Prototype of the system developed

In order to perform a set of experiments, a prototype rear-projection system was developed, consisting of a camera/projector pair, and a fine, flat, translucent projection surface. The process is controlled by a personal computer with an Intel Core 2 Duo processor, running at 2.66 GHz. The camera is a Logitech 9000 Pro webcam. The projector and the camera are located on the same side of the screen, which is the opposite side in relation to the user's location. The projector is aligned with the lower edge of the screen, but is pointing at its centre. The camera is located a little bit lower than the projector, so as not to interfere with its beam. The acquisition of images by the camera is not synchronized with the operations on the projector's frame buffer.

In this prototype, the correspondence between the estimated image pixels and the camera images may have an error of $\pm 1$ pixel. A filtering algorithm was designed to decrease the error in the detection of photometric perturbations. Once a binary segmented image has been obtained, a pixel with value one retains its value if, and only if, the sum of values in a $5 \times 5$ neighbourhood is greater than 10. Then, the resulting image is subjected to a morphological analysis that takes into account the area of the spots detected, and the concentration of points in those spots. The calculations for this analysis are kept as simple as possible, so as introduce the least possible overhead.

More precisely, we have considered two thresholds, $A_1$ and $A_2$, for the area of a spot, where $A_1 < A_2$. Spots with area less than $A_1$ are discarded outright, while those with area greater than $A_2$ are accepted. The remaining spots are accepted only if their area is greater than one fifth of the area of the minimal square that encloses all its points. Finally, we compute the centroids of the accepted spots.

## 9  Experimental results

The goal of experimentation was to confirm that the local photometric camera/projector calibration model provides greater reliability for detecting photometric perturbations than the global model.

In order to perform the comparisons, two applications were implemented for detecting photometric perturbations, both working on the prototype of the system described previously. One of the applications used the local photometric calibration model described in this work, and the other one used the global model described in Jaynes et al (2004). In order to monitor the detection of photometric perturbations, we used an auxiliary display, where the images taken by the camera were shown, with overlayed crosses to mark the centroids of the detected perturbations. Experimentation was divided into two phases: The first one considered only intrinsic photometric perturbations, while the second phase also included extrinsic photometric perturbations.

**Intrinsic photometric perturbations**

This experimentation phase is aimed at verifying the increase in reliability in presence of intrinsic photometric perturbations, when the local model is used, relative to existence of the spatial variations in the projected light intensity on the screen. Consequently, the experiments were carried out on a uniform monochromatic image, and photometric calibration was done in the absence of extrinsic photometric perturbations.

Figure 3 shows examples of images estimated by means of the local and global photometric calibration models. As it can be seen in Figure 3(a), the local model considers the existence of spatial variations of intensity in the light emitted by the projector upon the screen. On the other hand, as shown in Figure 3(b), the global model leads to a uniform image, thus discarding particular features of different regions of the screen.

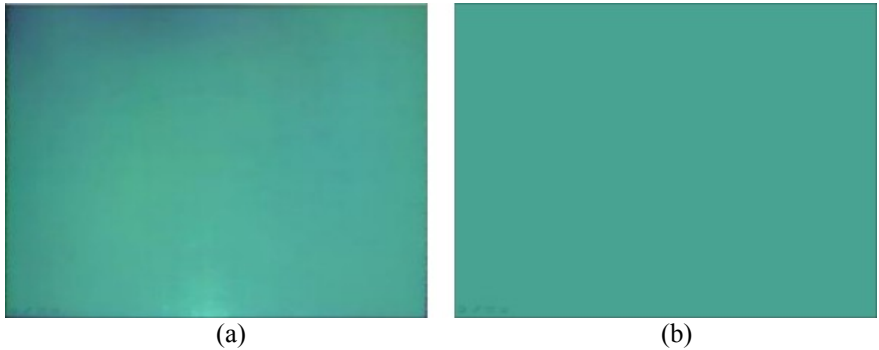

Figure 3. Image estimation by different photometric calibration models. (a) Image estimated under the local photometric calibration model. (b) Image estimated with the aid of the global photometric calibration model.

In order to carry out an adequate segmentation (according to Equation 8), we have computed for both models, and for each colour component (red, green, blue), the lower bounds for which photometric perturbations were correctly detected. For the local model, the lower bounds were (37, 25, 34), and for the global model we got the lower bounds (60, 62, 60).

Figure 4(a) shows a segmented image that has been obtained with the aid of the local model, for the bounds (37, 25, 34), while Figure 4(b) shows a segmented image obtained with the aid of the global model for the same thresholds. As it can be seen, the local model is robust under spatial variations of the intensity of the light projected on the screen, while the global model is not. This is due to the fact that in the local model, stationary perturbations are considered in the estimated images, and are therefore suppressed during image segmentation. That's why the local model gives smaller lower bounds than the global model.

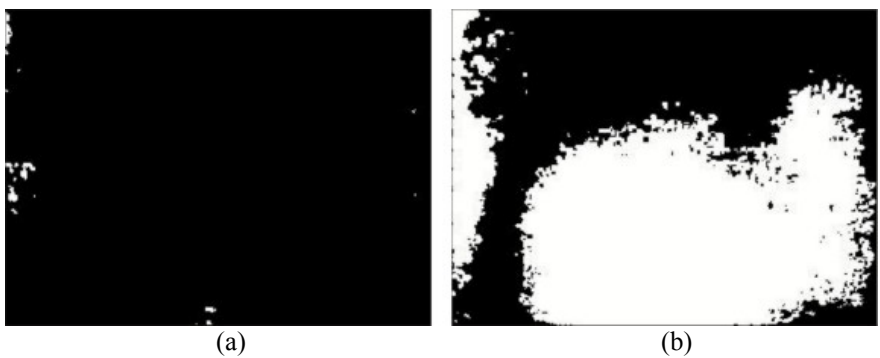

Figure 4. Segmented images for the lower bounds (37, 25, 34). (a) Image obtained with the aid of the local model. (b) Image obtained with the aid of the global model.

In order to create additional photometric perturbations, nine white rectangular objects were placed on the user's side of the screen during execution, after the photometric calibration phase. Figure 5(a) shows the results of a frame processed by means of the local model, for the threshold values (37, 25, 34). As it can be seen, the nine objects were correctly detected. However, these nice results were not obtained with the global model. Figure 5(b) shows the limits of the region where the objects were detected for the threshold values (60, 62, 60) with the global model.

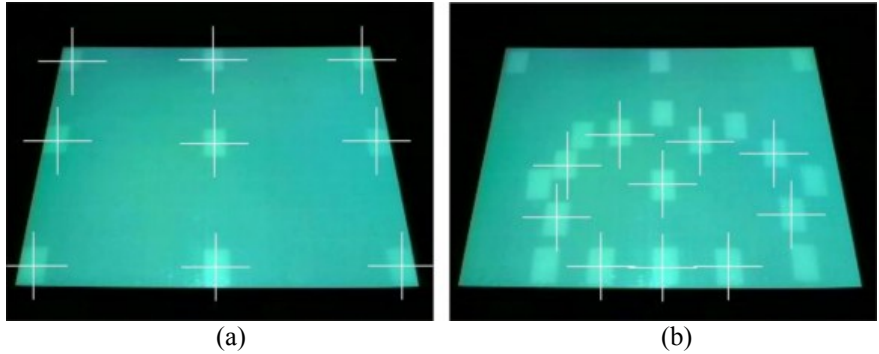

(a)  (b)

Figure 5. Results of object detection after photometric calibration. (a) Results obtained with the aid of the local model. (b) Results obtained with the aid of the global model.

In order to estimate a reliability measure of object detection, an object was dragged along the frontal side (user's side) of the screen, trying to cover as much space as possible, and we computed the percentage of frames where the object was correctly detected. The time interval for this experimentation was defined by pressing the buttons twice, once for the beginning, and one for the end. Table 1 shows the results obtained.

|  | % of detection |
|---|---|
| Application of the local model | 100% |
| Application of the global model | 41% |

Table 1. Percentage of reliability for detecting an object being dragged along the front side of the screen.

**Extrinsic photometric perturbations**

This experimentation phase is aimed at verifying the increase in reliability in presence of stationary photometric perturbations caused by external factors, when the local model is used. With that purpose, we created stationary photometric perturbations on the screen before performing the photometric calibration process, and these perturbations were maintained during all the experiments.

In the first kind of experiment, we placed nine white rectangular objects on the front side (user's side) of the screen. Figure 6(a) shows a frame taken by the camera, and Figure 5(b) shows an image estimated with the local photometric calibration model. As can be seen in Figure 6(b), the local model considers the existence of extrinsic stationary photometric perturbations on the screen. This is achieved by means of different estimation functions for each region of the screen. On the other hand, the global model obtains a uniform image, disregarding the particular features of each region of the screen.

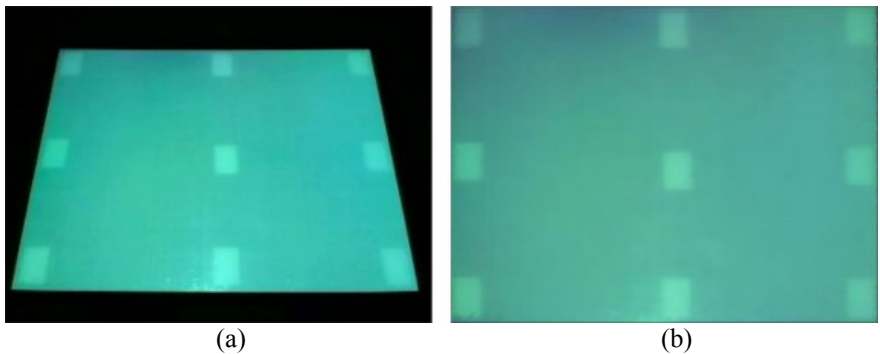

(a)                                                                                           (b)

Figure 6. Image estimation with the local photometric calibration model, under stationary photometric perturbations. Nine white rectangular regions have been placed on the screen, as external locally stationary photometric perturbations. (a) Uniform monochromatic image projected on the screen, and captured by the camera. (b) Image estimated by the local photometric calibration model.

Under the experimentation conditions described above, the lower bounds for correct detection of photometric perturbations were detected. For the local model, these bounds were (41, 29, 39), and for the global model they were (67, 83, 76). These values remained fixed during experimentation.

Figure 7(a) shows a frame processed with the local model, where ten extra objects were placed on the front side of the screen, after the photometric calibration process was carried out, thus creating new photometric perturbations. As can be seen, the ten non-stationary perturbations were detected, but none of the stationary perturbations was detected. Figure 7(b) shows the segmentation results for one frame, which describes the results better.

However, such nice results were impossible to get with the global model, where no photometric perturbation was detected. It is clear that if we lower the thresholds in order to detect non-stationary perturbations, stationary perturbations will be detected as well. Thus, the global model is definitely not suitable in the presence of stationary photometric perturbations.

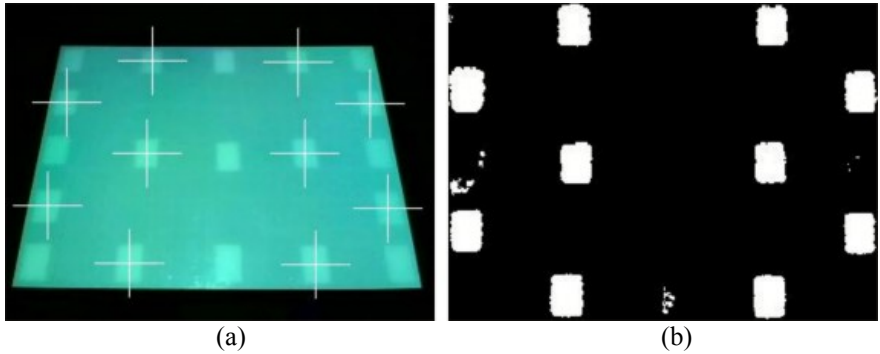

Figure 7. Frames processed with the local model, under stationary and non-stationary perturbations. (a) Frame. (b) Segmented frame.

When the experiments were performed on non-uniform images, the accuracy of the results decreased. Figure 8(a) shows the results of the experiments for a screenshot of the Windows XP explorer, where the effectiveness in detection of non-stationary perturbations was 100%. However, the same experiment was done with several landscape pictures, and in most cases there were non-stationary perturbations that were not detected. Figure 8(b) shows an example of that situation. On the average, the effectiveness of detection of non-stationary perturbations was around 77%.

In spite of the shortcomings, we regard these results as very satisfactory, for two main reasons: 1) in all cases, the non-stationary perturbations looked almost imperceptible to the human eye, and 2) none of the stationary perturbations was detected.

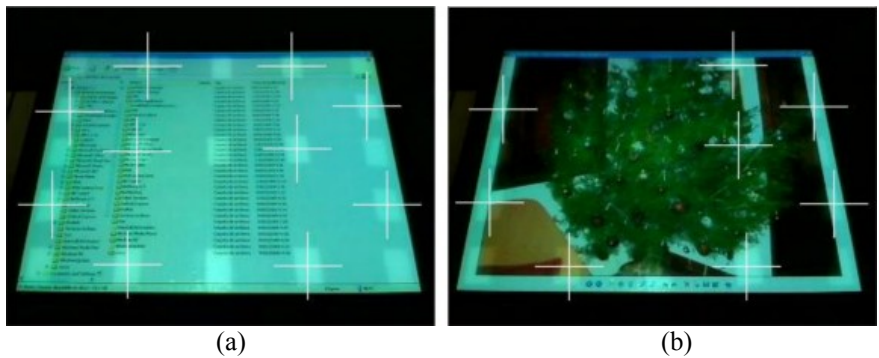

Figure 8. Frames showing non-uniform colour images, and subject to stationary and non-stationary perturbations. (a) 100% reliability in the detection of non-stationary perturbations (b) Some non-stationary perturbations were not detected.

When we experimented with very visible perturbations, the local method produced excellent results. Experiments were repeated several times, pointing a stationary flash light towards the screen, and moving two other flash lights around the screen. The effectiveness in detection was 100%. Figure 9(a) shows a frame with the incident light from the torch. Figure 9(b) shows how the two other non-stationary light spots, coming from respective torches, were detected.

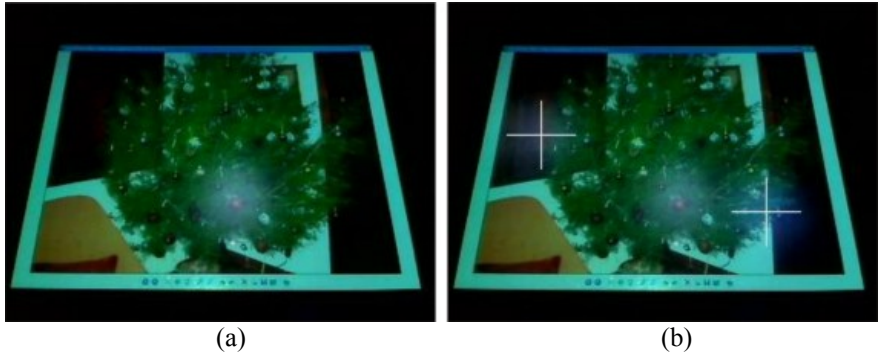

(a)                               (b)

Figure 9. Frames with very visible perturbations. (a) Frame with a flash light projecting a stationary light spot on the screen. (b) Detection of two non-stationary light spots created by two additional flash lights.

In short, the global photometric calibration model can be discarded when extrinsic locally stationary photometric perturbations are present, while the local model proved to be adequate in those circumstances. Table 2 gives a resume of the results obtained with the application of the local method.

| Type of perturbation | % of detection |
|---|---|
| Stationary | 0% |
| Slightly perceptible non-stationary | 77% |
| Highly perceptible non-stationary | 100% |

Table 2. Effectiveness of detection of stationary and non-stationary perturbations, with the use of the local model.

**Analysis of results**

From the experiments carried out in the presence of stationary perturbations, both intrinsic as well as extrinsic, we can conclude the following:
1. The application of the global photometric calibration model can be discarded in such circumstances.

2. The local model detected zero stationary photometric perturbations, either intrinsic or extrinsic.
3. The local model detected 100% of the non-stationary photometric perturbations that were very perceptible to the human eye.

The results obtained show that the method for detecting photometric perturbations in projected images, based on comparison between the projector's frame buffer and the corresponding camera buffer, with a local photometric calibration model, is robust under:

1. Locally stationary photometric perturbations, both intrinsic and extrinsic to the system.
2. Detection of highly visible non-stationary photometric perturbations.

## 10   Conclusions and future work

In this work we have presented a method for real-time detection of non-stationary photometric perturbations in projected images. The method is robust under locally stationary photometric perturbations, and therefore, to the existence of spatial variations of light intensity on the projection surface, to the influence of stationary external lights, and the existence of spots on the projection screen.

Our method seems to alleviate the shortcomings of other existing methods, such as Jaynes et al. (2004). However, in order to confirm the superiority of our method, we would need to design and carry out an extensive set of experiments, which is out of the scope of this work.

As future work, we think that it is also important to synchronize the acquisition of images between the camera, and the modifications in the projector's frame buffer, to decrease latency during the processing of frames, and improve the detection of slightly perceptible non-stationary photometric perturbations. As applications, we recommend the construction of rear tactile projection systems based on these results.

## References

AHLBORN, B.A., THOMPSON, D. and KREYLOS, O. (2005): A Practical System for Laser Pointer Interaction on Large Displays. *Proc. ACM VRST'05*, Monterey, California, USA: 106-109, ACM.

CASTAÑEDA-GARAY, M., BELMONTE-FERNÁNDEZ, O., GIL-ALTABA, J., PÉREZ-ROSÉS, H. and COMA, I. (2009): A Method for Real-time Detection of Photometric Perturbations in Projected Images (in Spanish). *Proc. CEIG 2009*, San Sebastian, Spain: 239-242, Eurographics.

ECKERT, R.R. and MOORE, J.A. (2000): The classroom of the 21st century: The interactive learning wall. *SIGCHI Bulletin*, **32**(2), 33-40.

JAYNES, C., WEBB, S. and STEELE, M. (2004): Camera-Based Detection and Removal of Shadows from Interactive Multiprojector Displays. *IEEE Transactions on Visualization and Computer Graphics* **10** (3): 290-301.

KIRSTEIN, C. and MULLER, H. (1998): Interaction with a projection screen using a camera-tracked laser pointer. *Proc. Int. Conf. MultiMedia Modeling*. IEEE Computer Society Press.

LAPOINTE, J.F. and GODIN, G. (2005): On-Screen Laser Spot Detection for Large Display Interaction. *Proc. HAVE'2005 (IEEE International Workshop on Haptic Audio Environments)*, Ottawa, Canada: 72-76, IEEE Computer Society Press.

OLSEN, D.R. and NIELSEN, T. (2001): Laser pointer interaction. *Proc. ACM Conf. Computer-Human Interaction* (CHI 2001), Seattle, WA, USA: 17-22, ACM.

REHG, J., FLAGG, M., CHAM, T.J., SUKTHANKAR, R. and SUKTHANKAR, G. (2002): Projected Light Displays Using Visual Feedback. Proc. ICARCV.

SUKTHANKAR, R., STOCKTON, R.G. and MULLIN, M.D. (2000): Self-Calibrating Camera-Assisted Presentation Interface. *Proc. Sixth Int. Conf. Control, Automation, Robotics and Vision (ICARCV)*, Singapore.

SUKTHANKAR, R., STOCKTON, R.G. and MULLIN, M.D. (2001): Smarter presentation: Exploiting homography in camera-projector systems. *Proc. Seventh Int. Conf. Control, Automation, Robotics and Vision (ICARCV)*, Vancouver, Canada: 247-253.

BRESNAHAN, G. et al. (2003): Building a large scale, high-resolution, tiled, rear projected, passive stereo display system based on commodity components. *Proc. Stereoscopic Displays and Virtual Reality Systems (SPIE X)*, **5006**: 19-30.

KRESSE, W., REINERS, D. and KNÖPFLE, C. (2003): Color consistency for digital multi-projector stereo display systems: The HEyeWall and the digital CAVE. *Proc. 9th Eurographics Workshop on Virtual Environments / 7th Immersive Projection Technologies Workshop*, Zurich, Switzerland: 271-279.

BROWN, M., MAJUMDER, A. and YANG, R. (2005): Camera-based calibration techniques for seamless multiprojector displays. *IEEE Trans. onVisualization and Computer Graphics* **11**: 193–206.

GONZALEZ, R., C., WOODS, R., E. (2002): Digital Image Processing (2$^{nd}$ ed). Prentice-Hall, Inc. Upper Saddle River, NJ.

FLYNN, M., J. (1972): Some computer organizations and their effectiveness. IEEE Trans. on Computers.